\documentclass{article}

\usepackage[nonatbib, preprint]{neurips}

\usepackage{float}

\usepackage[square,numbers]{natbib}
\usepackage[utf8]{inputenc} 
\usepackage[T1]{fontenc}    
\usepackage{hyperref}       
\usepackage{url}            
\usepackage{booktabs}       
\usepackage{amsfonts}       
\usepackage{nicefrac}       
\usepackage{microtype}      
\usepackage{graphicx}
\usepackage{tabularx}
\usepackage{hyperref}
\usepackage{amsmath,amssymb}
\usepackage{multirow}
\usepackage{subfigure}
\title{Enhanced Information Fusion Network for Crowd Counting}

\author{
  Geng Chen \\
  \texttt{gchenat@connect.ust.hk} \\
\And
  Peirong Guo \\
  \texttt{eeprguo@mail.scut.edu.cn} \\
}

\begin{document}
\maketitle

\begin{abstract}
  In recent years, crowd counting, a technique for predicting the number of people in an image, becomes a challenging task in computer vision. In this paper, we propose a cross-column feature fusion network to solve the problem of information redundancy in columns. We introduce the Information Fusion Module (IFM) which provides a channel for information flow to help different columns to obtain significant information from another column. Through this channel, different columns exchange information with each other and extract useful features from the other column to enhance key information. Hence, there is no need for columns to pay attention to all areas in the image. Each column can be responsible for different regions, thereby reducing the burden of each column. In experiments, the generalizability of our model is more robust and the results of transferring between different datasets acheive the comparable results with the state-of-the-art models.
\end{abstract}

\section{Introduction}
Crowd counting aims to predict the number of people in a given input image and generate the corresponding density map. This technology can be used for crowd density monitoring, analysis of crowd flow, traffic monitoring and so on. It can be applied extensively in security and commerce.

With the rise of deep learning, convolutional neural networks have also been applied to crowd counting. However, problems like the huge scale variation of crowd, the scenes and perspectives changes, bring great difficulties to crowd counting. To overcome such problems, an intuitive idea is to extract and fuse features at multiple levels. Some studies~\cite{zhang2016single,boominathan2016crowdnet,sam2017switching,cheng2019learning,cheng2019improving} have proposed the multi-column networks, which employ different sizes of convolution kernels in different columns, to capture features of various scales in the image. But this method has been pointed out that redundant information will be generated in multi-column networks. Therefore, the features obtained by different columns are often similar, and these structure can not overcome the variation of features. This will be the limiting factor for model performance. Therefore it is significant that different columns can extract different features in the image. An idea is to allow these columns to exchange information, so that the columns can obtain useful features from each other. From this perspective, we propose an Information Fusion Module (IFM) to build a channel for information interaction between columns. Considering that the generation process of the density map can be regarded as a gradual process, we introduce the intermediate supervision loss to refine the generated density map. This also allows the low-level parts of the network to notice the global information in the image. In experiments, we prove the effectiveness of this approach.

Secondly, the crowd counting needs to generate the corresponding density map. Making the position of the crowd in the generated density map close to the distribution of the ground truth has become another important issue. Shi et al.~\cite{shi2019counting} have introduced the idea of sementic segmentation into the crowd counting. In semantic segmentation, it needs to 
predict the category of each pixel. In crowd counting, pixels are simply classified as people or background. Since this is a binary classification problem, it is a natural choice to use sigmoid as the activation function of the output layer in the segmentation branch. However, the sigmoid function is smooth when the input's value is near to zero. Therefore, the model needs to predict a relatively small negative value in the background region and a relatively large positive value in the crowd region. On the other hand, the value of each pixel in the density map predicted is close to zero. This creates a conflict between the segmentation task and the counting task. Bai et al.~\cite{liao2020real} proposed a Differentiable Binarization (DB) module in text segmentation. To address the conflict, we modify this module's structure and further utilize a Steeper Differentiable Binarization (SDB) function to make it suitable for the segmentation task. This modification unifies the optimization target of crowd counting and segmentation and improves the quality of the prediction.

In summary, the major contributions of this paper are as follows: (1) The Information Fusion Module is proposed. It provides an information channel between different columns to reduce redundancy in the network. (2) We introduce the intermediate supervision loss for progressively generate more accurate density maps. (3) We introduce the Steeper Differentiable Binarization to unify the optimization target of segmentation task and counting task. The proposed IFNet shows strong transferability on different public benchmark datasets. And the quality of density maps generated by IFNet surpasses the state-of-the-art methods.

\section{Related Work}
\label{sec:related-word}

In the early studies, to solve the problem of crowd counting, detection methods were usually employed. These methods~\cite{dalal2005histograms,leibe2005pedestrian,viola2004robust,wu2007detection,felzenszwalb2009object} counted people by detecting the person or body part of the person in the image. They extracted the features of the image and then designed a detector to detect people. But these methods are only suitable for scenes with low crowd density. When the density gets higher, larger errors will occour. 

Therefore, to obtain better results in high-density situations, some regression-based methods~\cite{chan2009bayesian,chen2012feature,lempitsky2010learning} have been proposed. They extracted the low-level features of the image and the regressor mapped these features to the count number. However, in practical applications, it's important to get the density map, which reflects the distribution of the crowd.

As convolutional neural networks have exerted considerable influence on various fields of computer vision. It has also been introduced to the crowd counting. Due to the uneven distribution and irregular variation of crowd in different scenes, there are still many problems to be solved. In earlier research, CrowdNet~\cite{boominathan2016crowdnet} combined a deep column and a shallow column to extract different scale features. MCNN~\cite{zhang2016single} further proposed a three-column network. Various-size convolution kernels have been used in these columns to obtain multiple receptive fields, so that it tried to obtain multi-scale features. However, some researches found that different columns of MCNN cannot extract different features. To improve the performance of the model, Sam et al.~\cite{sam2017switching} designed a classifier to determine which column should be chosed to extract the features of the input image. Cheng et al.~\cite{cheng2019improving} estimated the mutual information between two columns and updated the parameters of these columns in turn. Liu et al.~\cite{liu2019crowd} integrated three subnetworks with different scaled inputs and got the density map by fusing multiple-level features of these subnetworks.

Some other methods aims to make the network better at the difficult parts of the input. Such as the attention mechanisms, these methods~\cite{liu2019adcrowdnet,sindagi2019inverse,zhang2019relational,bai2019crowd} allow the network to notice some important areas in input image.

Different from these methods, our model obtains information based on the response strength of signals between columns. The two columns exchange information through the Information Fusion Module, so that different columns can handle different information. And we also calculated the loss of the intermediate results. Such an intermediate supervision method enables the model to gradually make better predictions.

Studies~\cite{marsden2017resnetcrowd,sindagi2017cnn,idrees2018composition,liu2018decidenet} showed that learning multiple related tasks can reduce counting errors. An related task is to classify the images into different categories based on the number of people in the image. Designing a branch for classification is a common practice. This approach introduces additional information and helps the network training. However, there is a large gap in crowd density between different datasets. Therefore, the classification threshold is hard to decide and this approach may not have a positive impact on model's generalization performance. Shi et al.~\cite{shi2019counting} combined crowd counting, segmentation and classification tasks, which made the density map generated by the network more accurate. 

We also design a segmentation branch for predict the location of crowd. We introduce a Steep Differentiable Binarization (SDB) function and an adaptive bias term to compensate for this shortcoming of the sigmoid function. The experiments show it improves our model's performance.

\section{Method}
\label{sec:evaluation-tasks}
In this part, we introduce the Information Fusion Network (IFNet) and the optimization function. Our loss function integrates intermediate supervision loss and segmentation loss to make the model perform better. In particular, we will demonstrate the Information Fusion Module (IFM) in detail. Besides, we introduce the Steeper Differentiable Binarization (SDB) to help us unify the segmentation task and counting task.
\begin{figure}
  \centering
  \includegraphics[height=5cm]{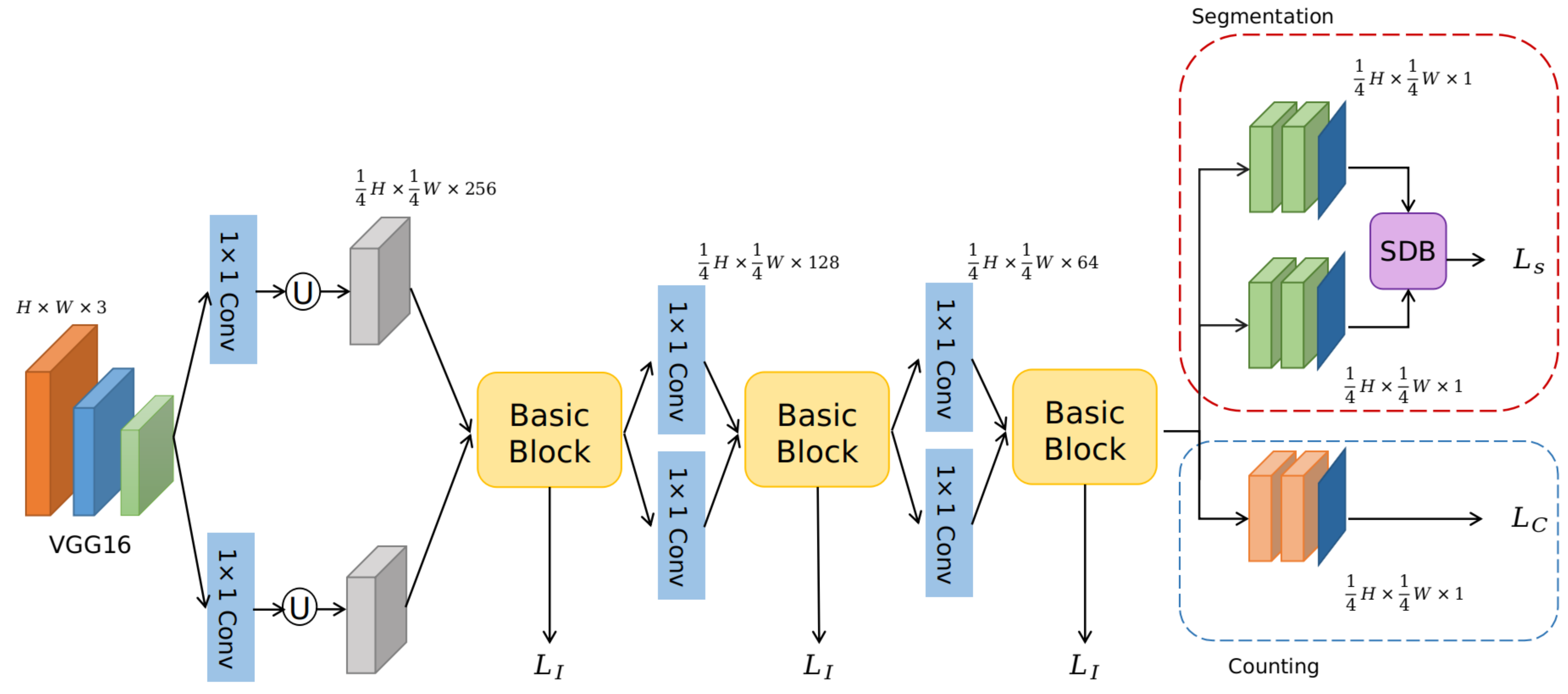}
  \caption{Information Fusion Network}                                       
  \label{fig:2}
\end{figure}
\subsection{Overview of IFNet}
In crowd counting, extracting multi-scale features is an important step. We introduce the Information Fusion Module (IFM) between columns. This enables two columns to process various regions in the image separately and can adaptively decide the information obtained from the other column to assist the prediction. Such columns do not need to be responsible for all regions, which reduces the burden of each column. Each column can obtain the information of a wider region according to the Information Fusion Module.

The overall pipeline of IFNet is shown in Figure 2. VGG-16~\cite{simonyan2014very} is chosen as backbone. But since the small feature maps will limit the performance, only the convolutional layers before the $5^{th}$ maxpooling layer will be kept and the shape of the output is $\frac{1}{8}H\times \frac{1}{8}W\times 512$, where $H$ and $W$ represent the height and width of the input image. The basic block's structure can be found in Figure. 3. It is similar to Hourglass\cite{newell2016stacked}, but it consists of two columns. These two columns send information to each other via an Information Fusion Module: 
\begin{align}
  F_{IF_{1}},F_{IF_{2}}=IFM(x_{1},x_{2})
\end{align}
where $ IFM$ represents the Information Fusion Module, $F_{IF_{i}}$ represents the output of Information Fusion Module and $x_{i}$ represents the corresponding input. Then the $F_{IF_{i}}$ will provides $x_{i}$ with important information to enhance the training process:
\begin{align}
  F_{out_{i}}=F_{IF_{i}}+x_{i}
\end{align}
This structure makes our network easier to map features to density maps. The outputs of these columns are finally integrated to generate a density map, which is used for calculating the intermediate supervision loss $L_{I}$. This density map is then added to the output of the basic block by increasing the number of channels through a $1\times 1$ convolution kernel.

IFNet has three branches for prediction and the shape of outputs are $\frac{1}{4}H\times \frac{1}{4}W\times 1$. Counting branch is responsible for counting task, and the other two branches combine with Steeper Differentiable Binarization are responsible for segmentation task. In crowd counting, our target is to distinguish the crowd from the background in the image. Obviously, the segmentation branches predict the position of crowd, which introduces additional auxiliary information for our counting task.
\begin{figure}
  \centering
  \includegraphics[height=5cm]{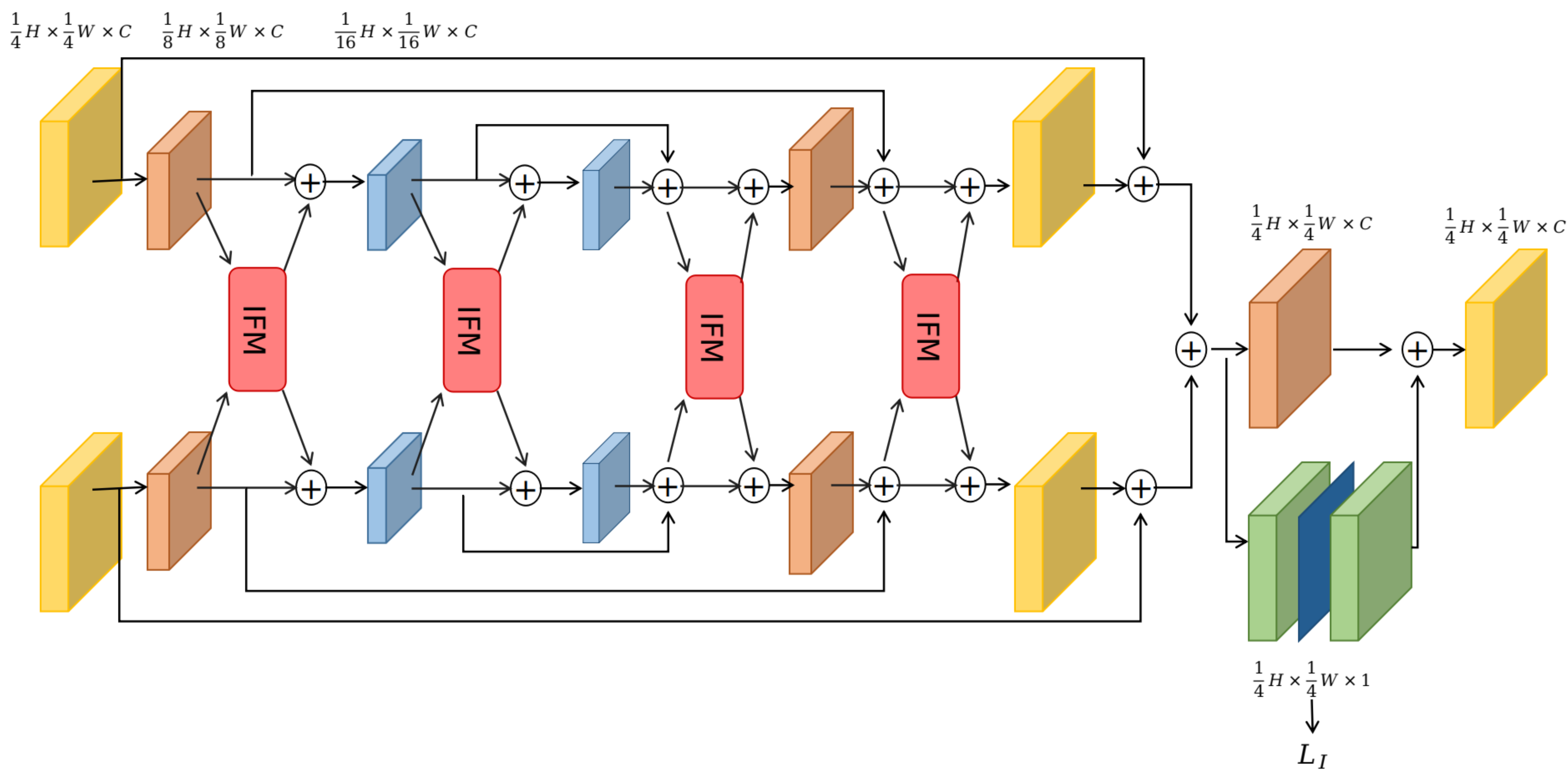}
  \caption{Basic Block}
  \label{fig:3}
\end{figure}
\subsection{Information Fusion Module}
\begin{figure}
  \centering
  \includegraphics[height=5.5cm]{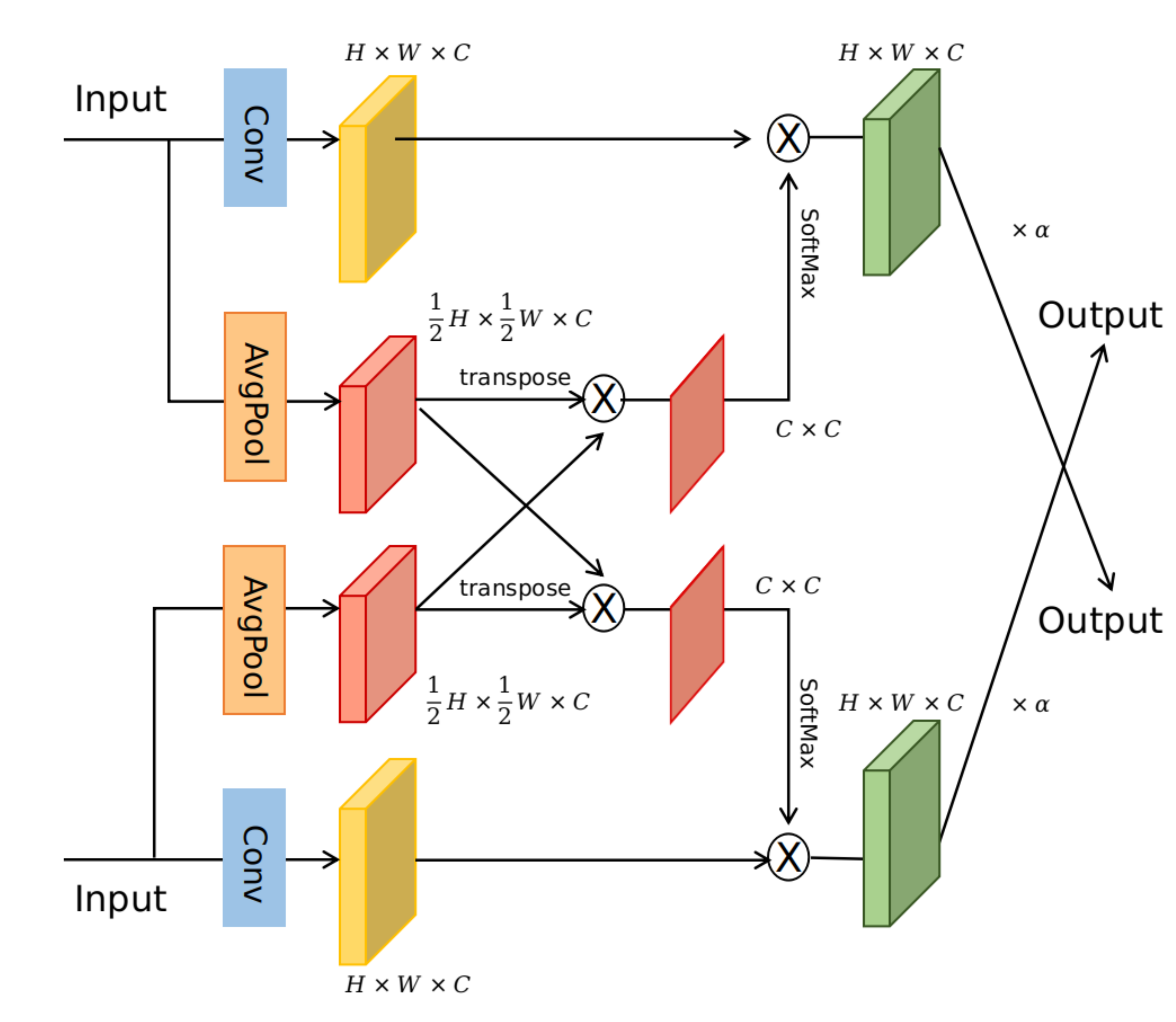}
  \caption{Information Fusion Module (IFM)}
  \label{fig:4}
\end{figure}
As illustrated in Figure 4, this module is designed for information exchange between two columns. First, for the two inputs $x_{1}$ and $x_{2}$ of shape $H\times W\times C$, we pass them through different $3\times3$ convolution kernels to extract the feature $F_{C_{i}}$.

Since the points in the density map have a large correlation in a small region, we use Average pooling to condense the input features to get $D_{i}$. The shape of $D_{i}$ is $\frac{1}{2}H\times \frac{1}{2}W\times C$. We reshape and transpose then to $\mathbb{R}^{C\times HW}$, which means that each row represents a feature map of the original features. Then we calculate the signal intensity $G\in \mathbb{R}^{C\times C}$:
\begin{align}
  G_{i}=D_{i}D^{T}_{j}
\end{align}
The $j^{th}$ element in the $i^{th}$ row of G represents the response of the $i^{th}$ channel of the original feature map in one column to the $j^{th}$ channel of the original feature map in the other column. Then we normalize each row to get $W\in \mathbb{R}^{C\times C}$:
\begin{align}
  w_{i,j}=\frac{exp(g_{i,j})}{\sum_{j}^{C}exp(g_{i,j})}
\end{align}
where $w_{i,j}$ is the response of one column's $i^{th}$ channel to the other column's $j^{th}$ channel. Finally, we multiply $F_{C}$ with $W$. And we set a parameter $\alpha$ to scale these information:
\begin{align}
  F_{IF_{i}}=\alpha\times W_{j}F_{C_{j}}
\end{align}

\subsection{Steeper Differentiable Binarization}
Because the sigmoid function changes slowly from zero to one, it is necessary to keep the input of the sigmoid function away from zero when the network completes the segmentation task. In contrast, for counting task, many values of the density map are close to zero, which causes a conflict between these two tasks. So we try to employ a Steeper Differentiable Binarization (SDB) function and an adaptive bias term to solve this problem. The Steeper Differentiable Binarization is defined as:
\begin{align}
  M_{i,j}=\frac{1}{1+exp(-k(P_{i,j}-B_{i,j}))}
\end{align}
where $M$ is the prediction mask of segmentation, $P$ and $B$ are the outputs of the segmentation branches. BCE loss is the optimization function for segmentation task:
\begin{align}
  L_{S}=\frac{1}{N}\sum_{i=1}^{N} y_{i}log(m_{i})+(1-y_{i})log(1-m_{i})
\end{align}
where $y_{i}$ represents the ground truth, $m_{i}$ is our prediction and $N$ is number of training samples. Define $t=P_{i,j}-B_{i,j}$, then the differentia of $L_{s}$ is:
\begin{align}
  L_{S}^{'}=\frac{\partial L}{\partial t}=\frac{1}{N}\sum_{i=1}^{N}k(y_{i}-\frac{1}{1+exp(-kt)})
\end{align}
It can be seen that the factor $k$ will affect the gradient directly. Therefore, in next section, we adjust it by a parameter $\lambda$.
\subsection{Loss Function}
Our loss function consists of three parts: MSE loss for the intermediate density maps $L_{I}$, MSE loss for final prediction of density map $L_{C}$ and BCE loss for segmentation task $L_{S}$. $L_{I}$ helps the model generate some initial predictions of density maps. It can be seen as a refining process that gradually produces more accurate density maps. $L_{S}$ provides auxiliary location information to IFNet and $L_{C}$ is the main target of our network.  

Therefore, the total loss function $L$ is given by:
\begin{align}
  L=\frac{\alpha}{4}(L_{I}+L_{C})+\lambda L_{S}
\end{align}
where $\alpha = 1$ and $\lambda = 0.005$. Because the Steeper Differentiable Binarization function will make the gradient values of segmentation branch large, we set $\lambda$ to a small value.
\section{Experiment}
\label{sec:methods}

\subsection{Datasets}
\textbf{ShanghaiTech}~\cite{zhang2016single} has a total of 1198 images and 330,165 annotated people. It is divided into two parts. One is Part A with 482 images, which contains densely crowds (the number of people ranging from 33 to 3139). The other is Part B with 716 images, where crowds are sparse (the number of people ranging from 9 to 578).\\
\textbf{UCF\_CC\_50}~\cite{idrees2013multi} contains 50 crowd images with different densities. Because these images have different resolutions and various densities and the dataset is relatively small, it becomes a challenging dataset in crowd counting.\\
\textbf{UCF-QNRF}~\cite{idrees2018composition} is a recently proposed dataset that contains 1535 images with different resolutions and the number of people ranging from 49 to 12865. Its images come from various scenes in real world, which can reflect the requirement of practical applications and have more experimental value.

\subsection{Evaluation Metrics}
Consistent with standard evaluation metrics, we employ MAE and MSE to measure the performance of our model. MAE and MSE are defined as $MAE=\frac{1}{N}\sum_{i=1}^{N}\left | C_{i}-C_{i}^{'} \right |$ and $MSE=\sqrt{\frac{1}{N}\sum_{i=1}^{N}\left | C_{i}-C_{i}^{'} \right |^{2}}$, where $C_{i}$ is the predict result of $i^{th}$ image, $C_{i}^{'}$ denotes the corresponding ground truth count and $N$ is number of test images.
\subsection{Training settings}
As in~\cite{gao2019c}, we choose a fixed-size Gaussian kernel to generate the density maps. Then we resize the images and the corresponding density maps so that its width and height can be divided by 16 and not greater than 1024.

During training, we random crop patches of size [224, 224] from the images and the batchsize is set to 16. The parameters of the IFNet are initialized by Xavier~\cite{glorot2010understanding}. Adam~\cite{kingma2014adam} optimizer with the learning rate of $5\times 10^{-5}$ is used to train our model and the learning rate is halved every 1000 epochs. We random flip images horizontally and scale brightness and saturation for data augmentation. When inference, image will be fed into our model directly.
\subsection{Ablation Study}
We perform ablation study on the ShanghaiTech Part A and UCF-QNRF datasets, which contain some challenging samples.
\begin{figure}
  \centering
  \begin{minipage}{0.23\linewidth}
		\subfigure{\begin{minipage}[t]{0.2\linewidth}
				\centering
				\includegraphics[width=2.85cm]{A23}
		\end{minipage}}
  \end{minipage}
  \begin{minipage}{0.7\linewidth}
  \subfigure{
  \begin{minipage}[t]{0.13\linewidth}
  \centering
  \includegraphics[width=1.4cm]{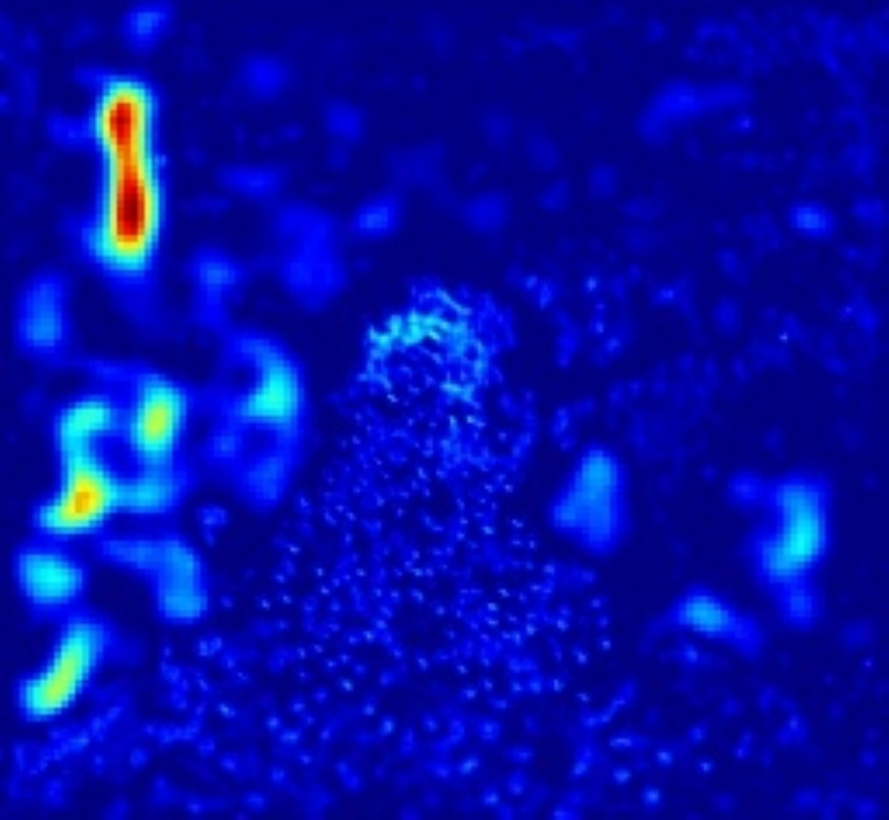}
  \includegraphics[width=1.4cm]{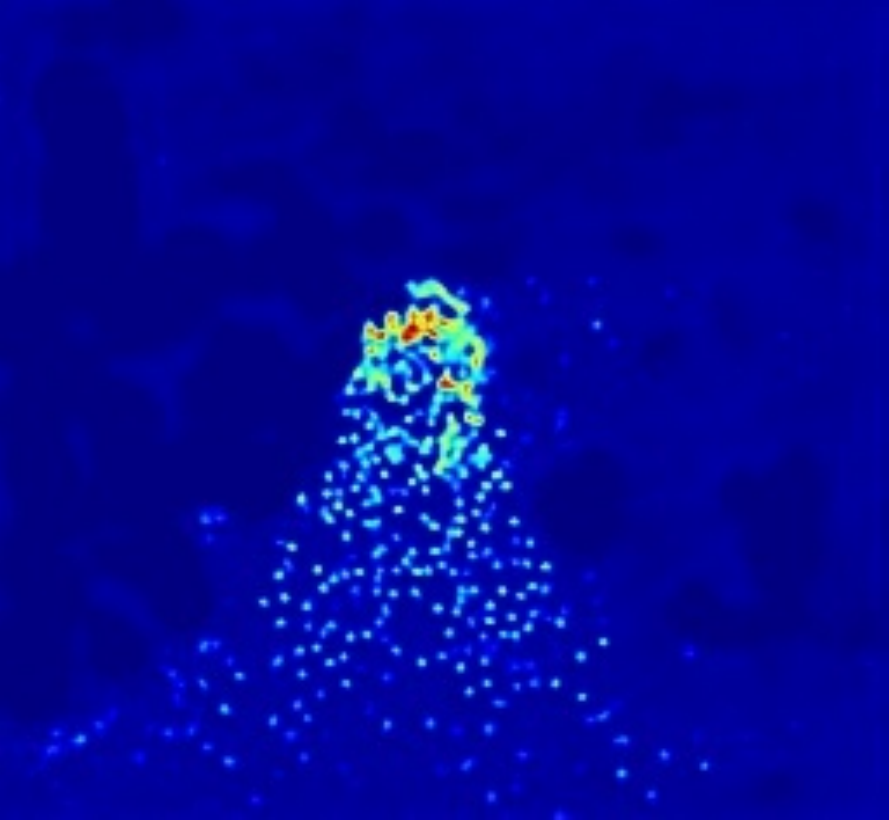}
  \end{minipage}
  }
  \subfigure{
  \begin{minipage}[t]{0.13\linewidth}
  \centering
  \includegraphics[width=1.4cm]{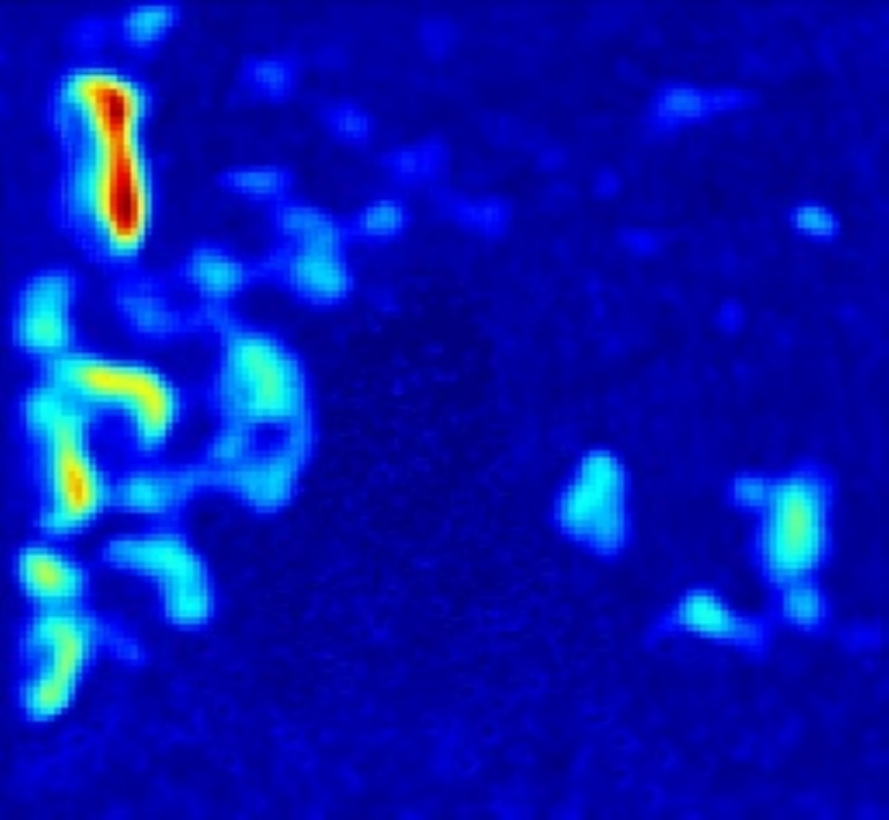}
  \includegraphics[width=1.4cm]{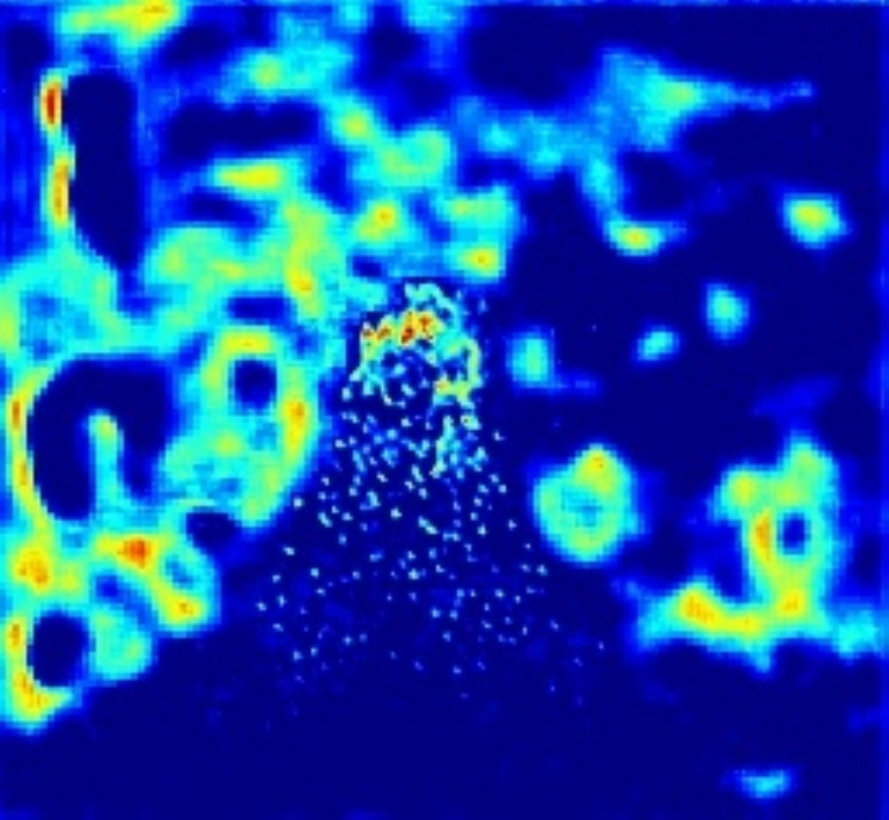}
  \end{minipage}
  }
  \subfigure{
  \begin{minipage}[t]{0.13\linewidth}
  \centering
  \includegraphics[width=1.4cm]{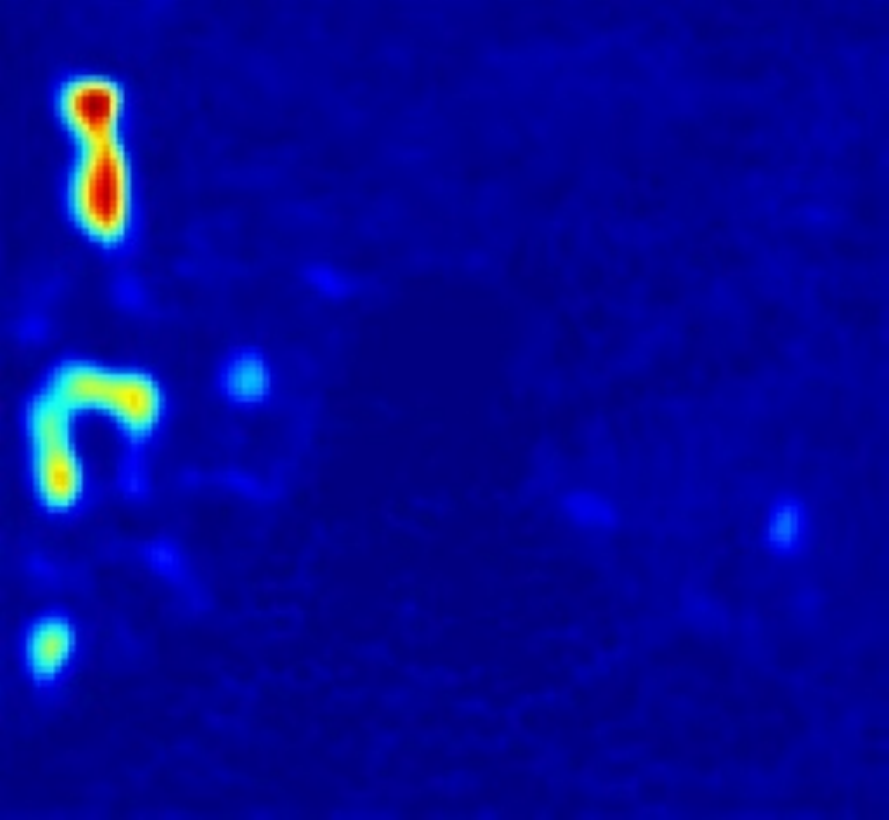}
  \includegraphics[width=1.4cm]{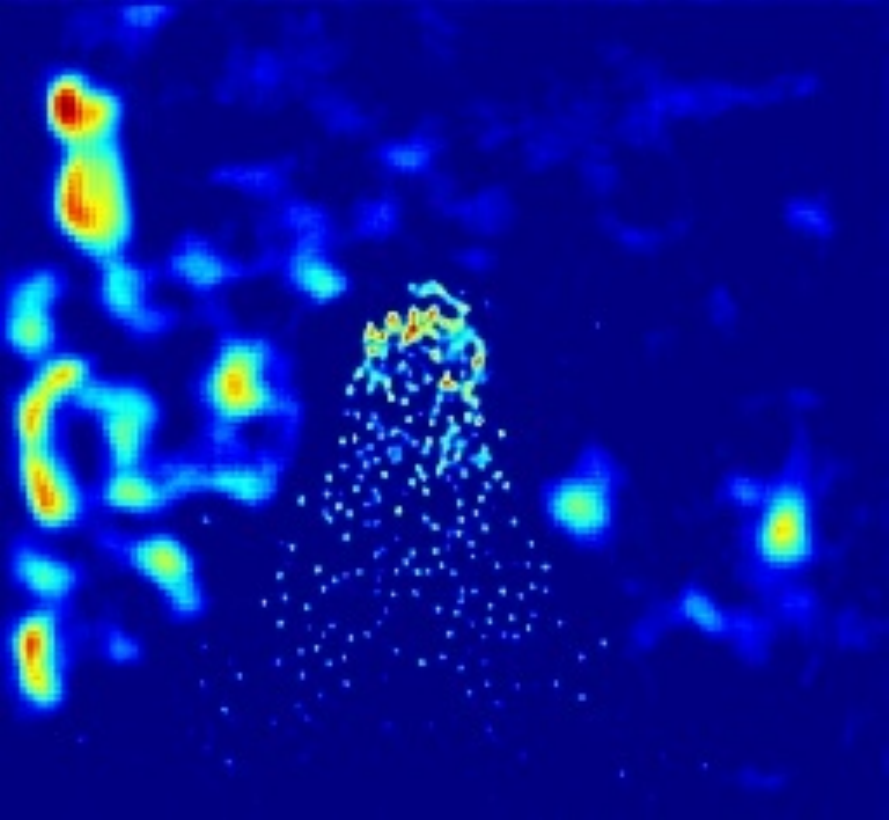}
  \end{minipage}
  }
  \subfigure{
  \begin{minipage}[t]{0.13\linewidth}
  \centering
  \includegraphics[width=1.4cm]{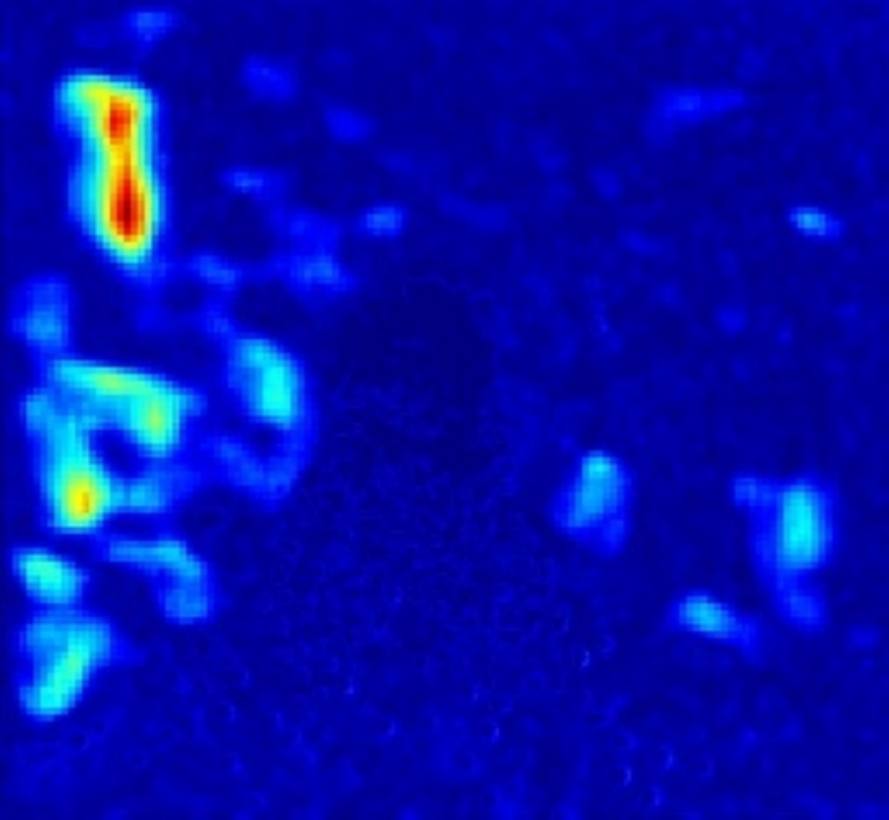}
  \includegraphics[width=1.4cm]{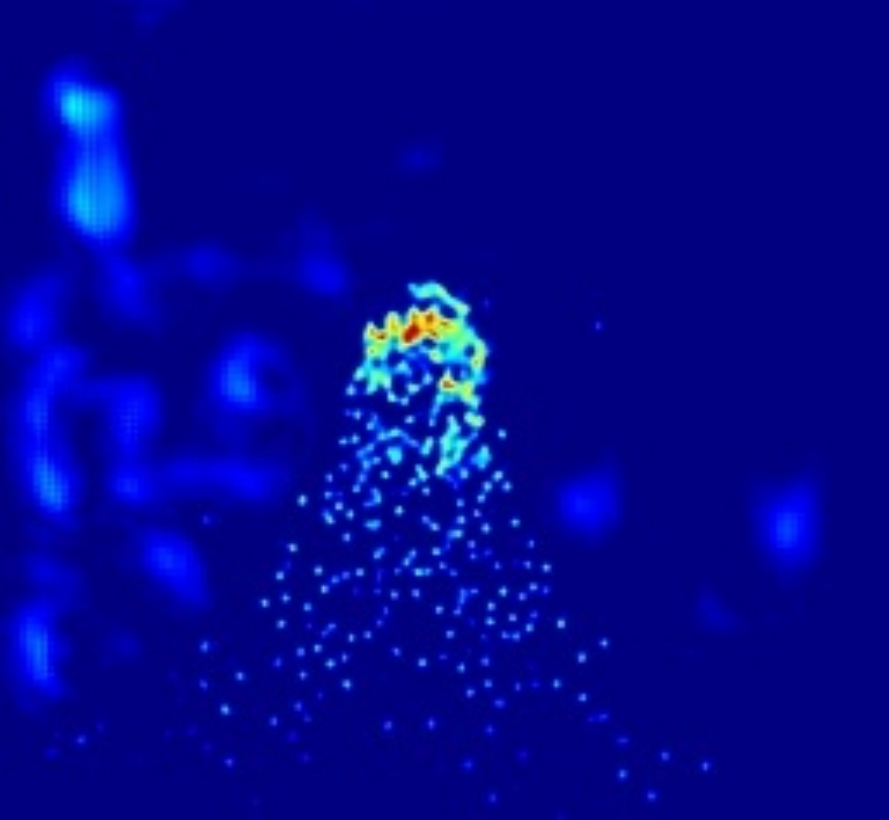}
  \end{minipage}
  }
  \subfigure{
  \begin{minipage}[t]{0.13\linewidth}
  \centering
  \includegraphics[width=1.4cm]{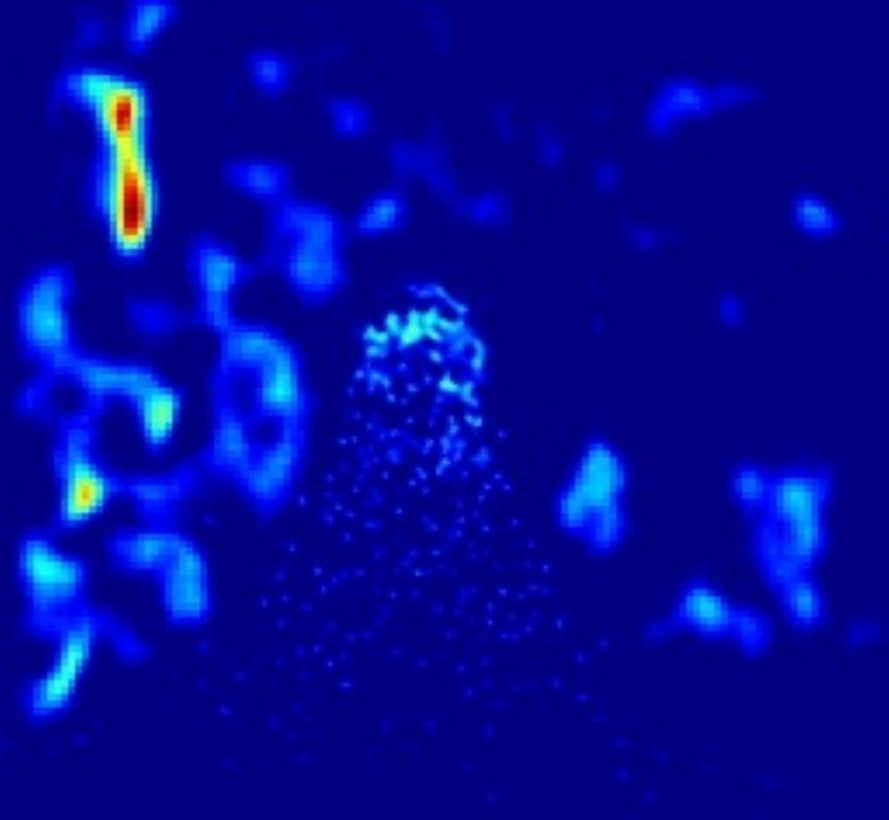}
  \includegraphics[width=1.4cm]{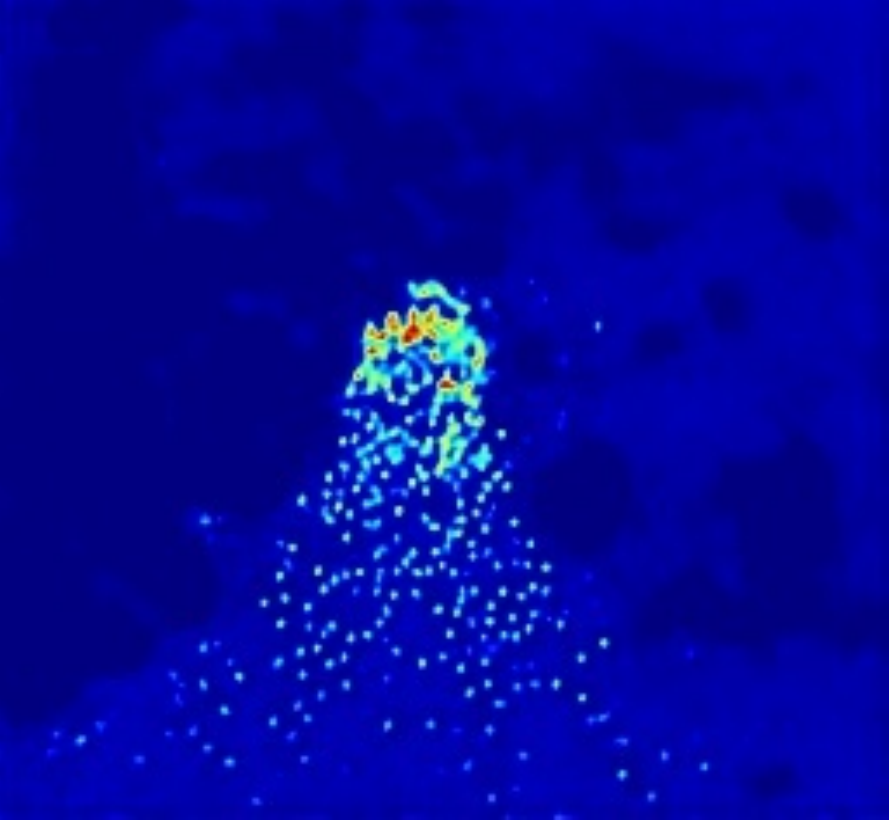}
  \end{minipage}
}
\subfigure{
  \begin{minipage}[t]{0.13\linewidth}
  \centering
  \includegraphics[width=1.4cm]{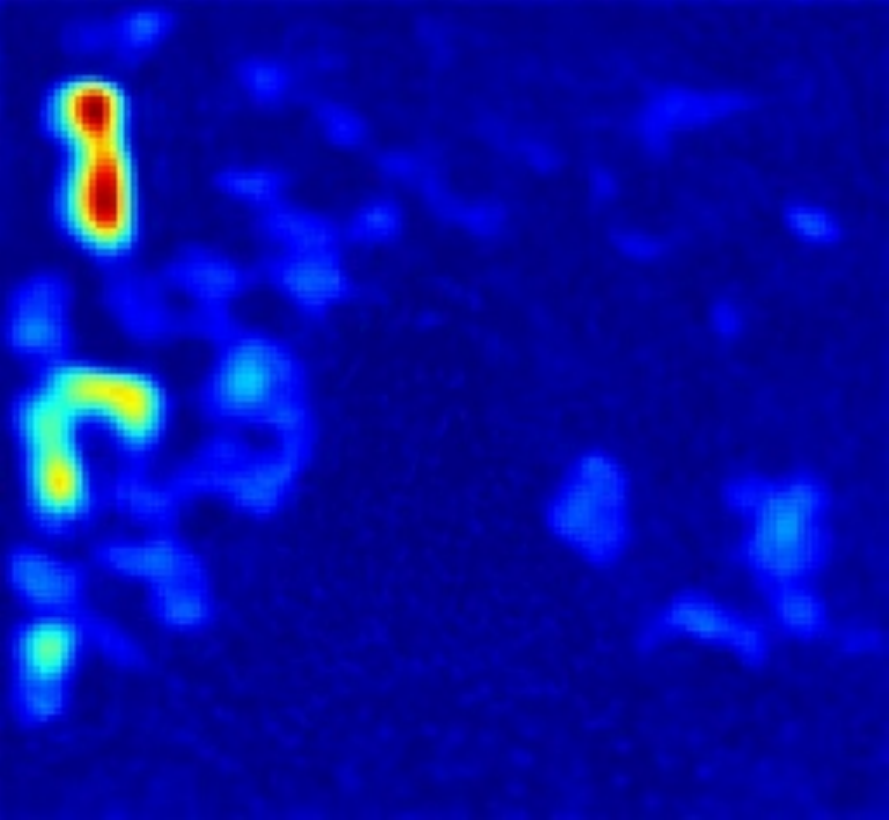}
  \includegraphics[width=1.4cm]{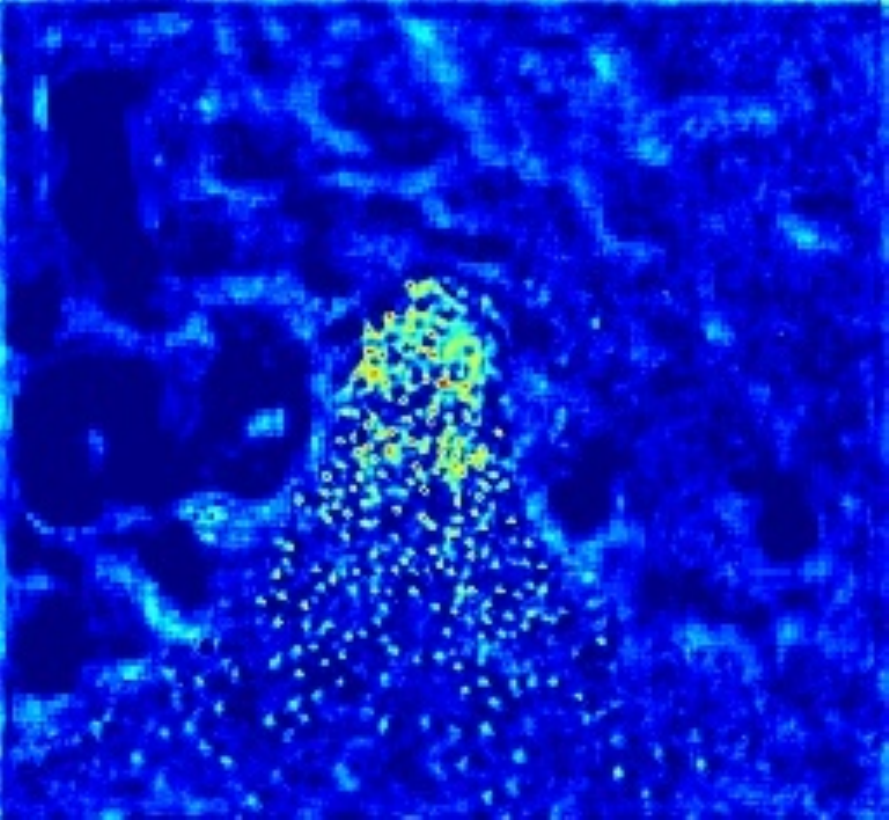}
  \end{minipage}
}

\end{minipage}

  \caption{Visualization of feature maps generated by different columns. First two rows: RGB images, the feature maps generated by one column. Last two rows: ground truth, the feature maps generated by another column}
  \label{fig:5}
\end{figure}

\textbf{Information Fusion Module.} The Information Fusion Module provides a channel for information communication between two columns, so that columns can focus on different regions of the image. As mentioned above, we use $\alpha$ for scaling the information and its effects are shown in the Table 1. It can be seen from the results that when $\alpha$ is too large, the performance is even worse than the model without the Information Fusion Module. The reason is that if $\alpha$ is too large, some features will be greatly enhanced, which will weakens the model's response to other features. When the Information Fusion Module is not employed, these two columns cannot interact with each other, which results in redundant information in the network. To demonstrate our module enables columns to extract multi-scale features, we visualize the feature maps of the two columns and the visualization results are shown in Figure 4. It points out that the first column will prefer to focus on background and the low-density regions. In contrast, the other column pays more attention to the high-density regions. It proves that different columns of IFNet can extract different features and overcome the scale variation.
\begin{table}
\label{table:Ablation study}
\centering
\begin{minipage}[t]{0.45\textwidth}
   \centering
   \caption{Evaluation of Information Fusion Module}
   \scriptsize
   \begin{tabular}{|c|c|c|c|c|}
   \hline
     & \multicolumn{2}{c|}{ShanghaiTech Part A} & \multicolumn{2}{c|}{UCF-QNRF}\\ \cline{2-5} & MAE & MSE & MAE & MSE \\
   \hline\hline
   w/o IF & 64.6 & 113.7 & 104.9 & 187.5\\
   $\alpha$ = 1 & 75.2 & 131.3 & 113.0 & 209.6\\
   $\alpha$ = 0.5 & 63.3 & 112.5 & 106.8 & 196.1\\
   \hline\hline
   $\alpha$ = 0.3 & \textbf{61.1} & \textbf{111.7} & \textbf{98.2} & \textbf{184.6}\\
   \hline
   \end{tabular}
\end{minipage}
\begin{minipage}[t]{0.45\textwidth}
   \label{table:Ablation study}
   \centering
   \caption{Evaluation of Steeper Differentiable Binarization}
   \scriptsize
   \begin{tabular}{|c|c|c|c|c|}
   \hline
     & \multicolumn{2}{c|}{ShanghaiTech Part A} & \multicolumn{2}{c|}{UCF-QNRF}\\ \cline{2-5} & MAE & MSE & MAE & MSE \\
   \hline\hline
   w/o SDB & 65.2 & 115.6 & 104.2 & 196.8 \\
  $k$ = 1 & 65.0 & 118.4 & 102.3 & 191.8\\
  $k$ = 50& 63.5 & 116.7 & 102.9 & 188.9 \\
  \hline\hline
  $k$ = 500 & \textbf{61.1} & \textbf{111.7} & \textbf{98.2} & \textbf{184.6}\\
   \hline
   \end{tabular}
\end{minipage}
\end{table}

\textbf{Steeper Differentiable Binarization.} The segmentation branch introduces additional location information to our task and assists the model to better locate location of crowd. However, the slow-changing character of the sigmoid function causes conflict between segmentation task and counting task. We introduce Steeper Differentiable Binarization to solve this problem. Table 2 shows the results of various settings. When $k$ is small, the binarization function is relatively flat. Therefore the input of the activation function should have large positive values in crowd regions and small negative values in other regions. This will negatively affect to the counting branch. But when $k$ is large, the activation function becomes steep. The segmentation branch only need to predict small values to get the binarized result. This approach unifies the optimization target of these tasks.

\textbf{Intermediate Supervision.} Each basic block in our model generates a density map, which is used for calculating the intermediate supervision loss. In this way, the low-level basic block can also capture the global features of the image. Through the cascading of these basic blocks, the density map will be continuously refined to achieve more accurate result. In order to verify its effect, we remove the sub-branches that generate the density map in the basic block (w/o IS branch) and discard the density map generated by the basic block to calculate the loss (w/o IS loss). As shown in Table 3, the intermediate supervision obviously improves the model's performance. Using this method can make the model pay attention to global and local features as much as possible at each stage.
\begin{table}
   \label{table:Ablation study}
   \centering
   \begin{minipage}[t]{0.45\textwidth}
      \centering
         \caption{Evaluation of Intermediate Supervision}
         \label{table:transferability}
         \scriptsize
         \begin{tabular}{|c|c|c|c|c|}
         \hline
         \multirow{2}*{Method}  & \multicolumn{2}{c|}{Part A} & \multicolumn{2}{c|}{UCF-QNRF}\\ \cline{2-5} & MAE & MSE & MAE & MSE\\
         \hline\hline
         w/o IS branch & 65.9 & 118.3 & 101.7 & 190.1 \\
         w/o IS loss & 65.7 & 117.7 & 104.8 & 185.9\\
         \hline\hline
         \textbf{IFNet (ours)} & \textbf{61.1} & \textbf{111.7} & \textbf{98.2} & \textbf{184.6}\\
         \hline
         \end{tabular}
   \end{minipage}
   \begin{minipage}[t]{0.45\textwidth}
      \centering
         \caption{Comparison of the density map quality}
         \label{table:transferability}
         \scriptsize
         \begin{tabular}{|c|c|c|c|c|c|c|}
         \hline
         \multirow{2}*{Method}  & \multicolumn{2}{c|}{Part A} & \multicolumn{2}{c|}{Part B} \\ \cline{2-5} & PSNR & SSIM & PSNR & SSIM\\
         \hline\hline
         MCNN~\cite{zhang2016single} & 21.4 & 0.52 & - & - \\
         CP-CNN~\cite{sindagi2017generating} & 21.7 & 0.72 & - & - \\
         CSRNet~\cite{li2018csrnet} & 23.8 & 0.76 & 27.0 & 0.89 \\
         ANF~\cite{zhang2019attentional} & 24.1 & 0.78 & - & - \\
         \hline\hline
         \textbf{IFNet (ours)} & \textbf{28.5} & \textbf{0.83} & \textbf{34.8} & \textbf{0.94} \\
         \hline
         \end{tabular}
   \end{minipage}
\end{table}
\begin{figure}
  \centering
  \subfigure{
  \begin{minipage}[t]{0.2\linewidth}
  \centering
  \includegraphics[width=2.7cm]{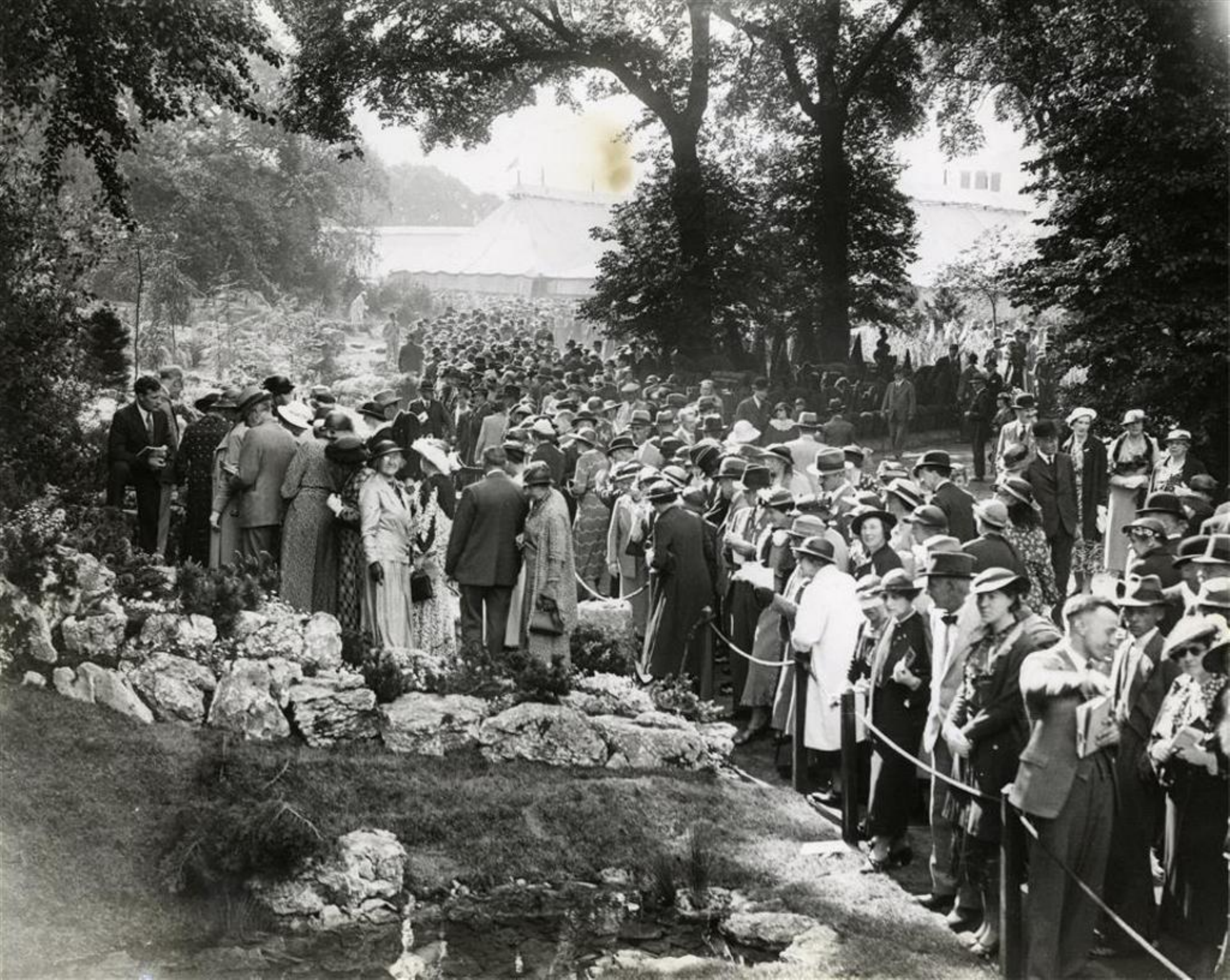}
  \includegraphics[width=2.7cm]{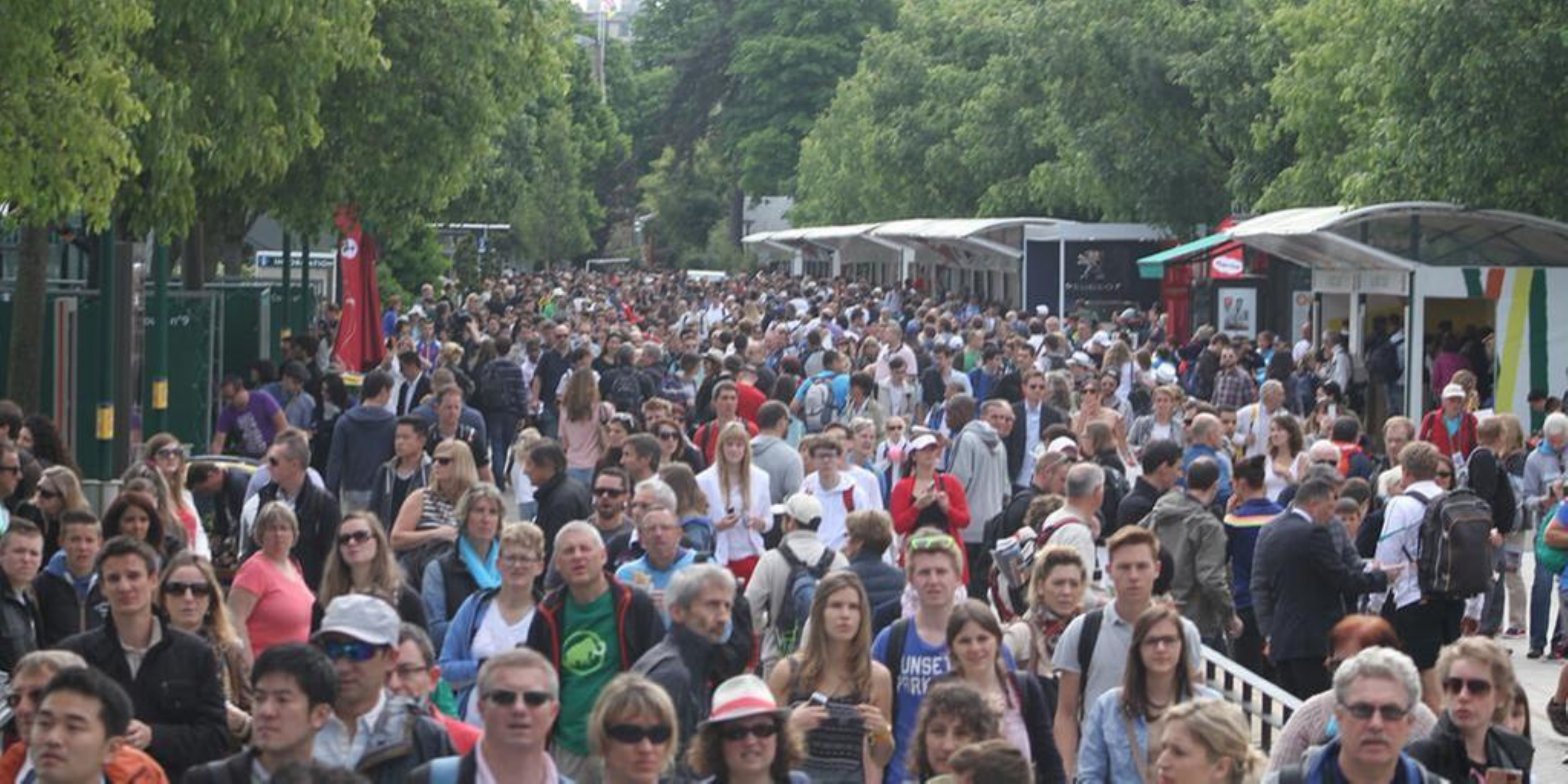}
  \end{minipage}
  }
  \subfigure{
  \begin{minipage}[t]{0.2\linewidth}
  \centering
  \includegraphics[width=2.7cm]{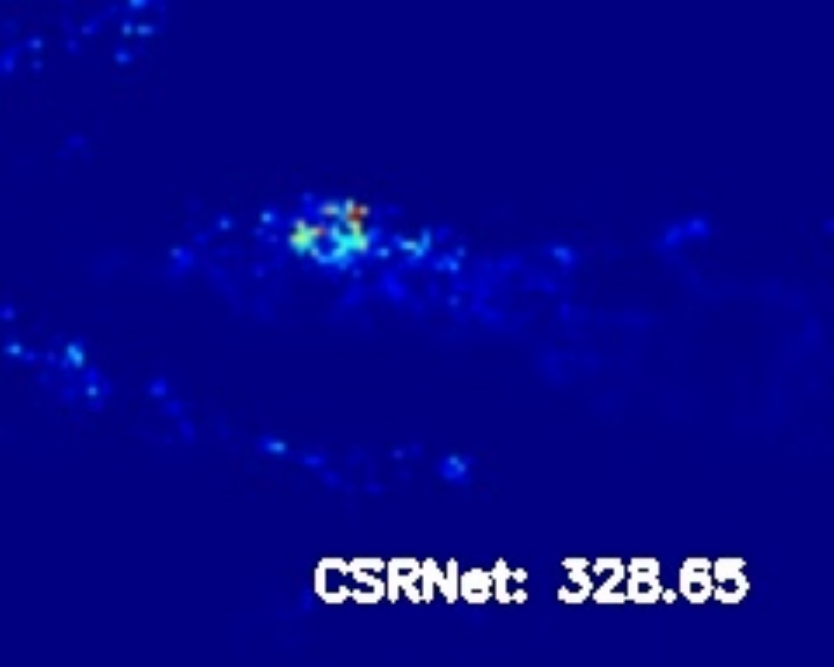}
  \includegraphics[width=2.7cm]{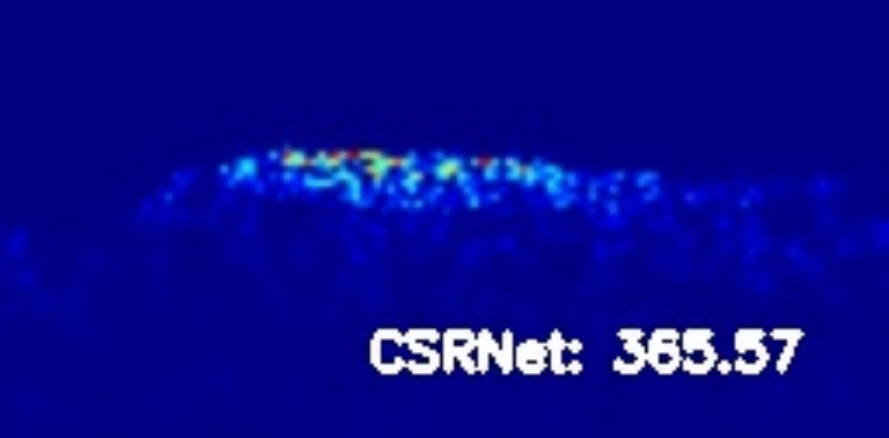}
  \end{minipage}
  }
  \subfigure{
  \begin{minipage}[t]{0.2\linewidth}
  \centering
  \includegraphics[width=2.7cm]{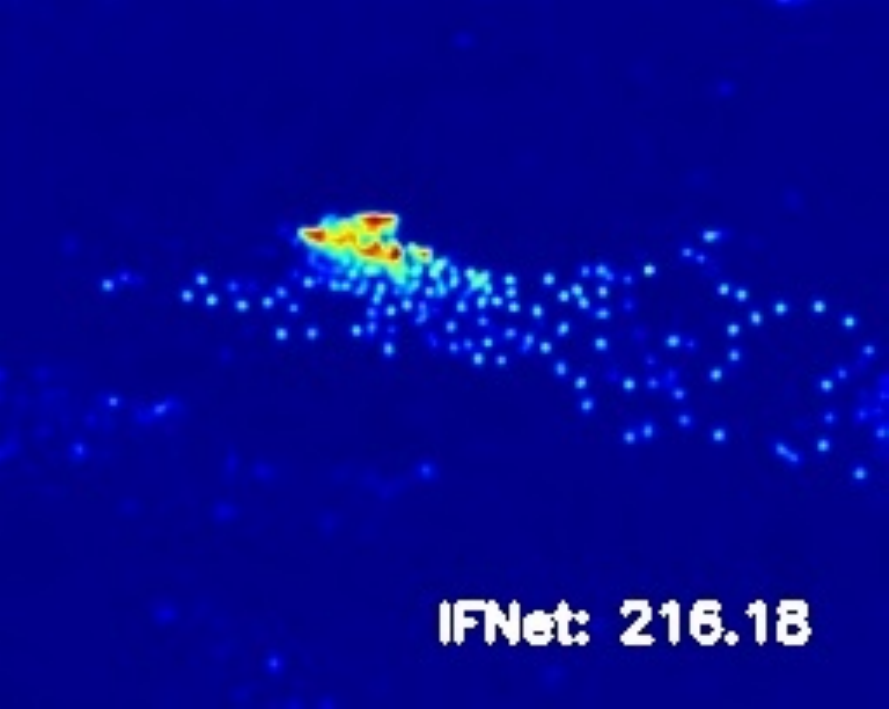}
  \includegraphics[width=2.7cm]{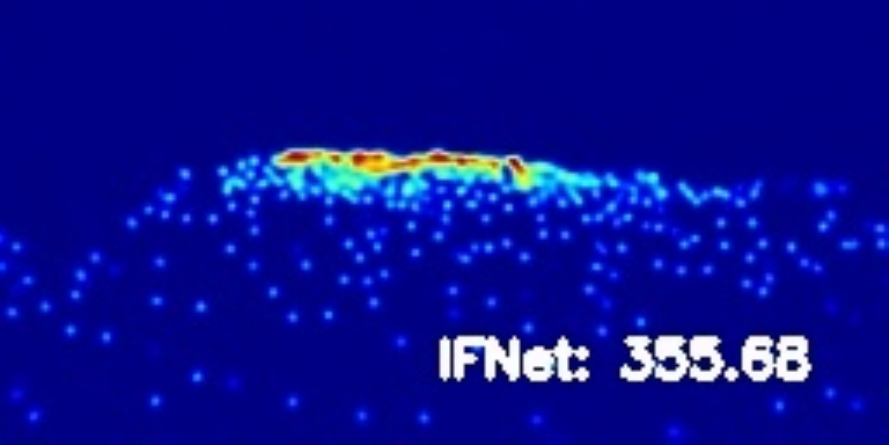}
  \end{minipage}
  }
  \subfigure{
  \begin{minipage}[t]{0.2\linewidth}
  \centering
  \includegraphics[width=2.7cm]{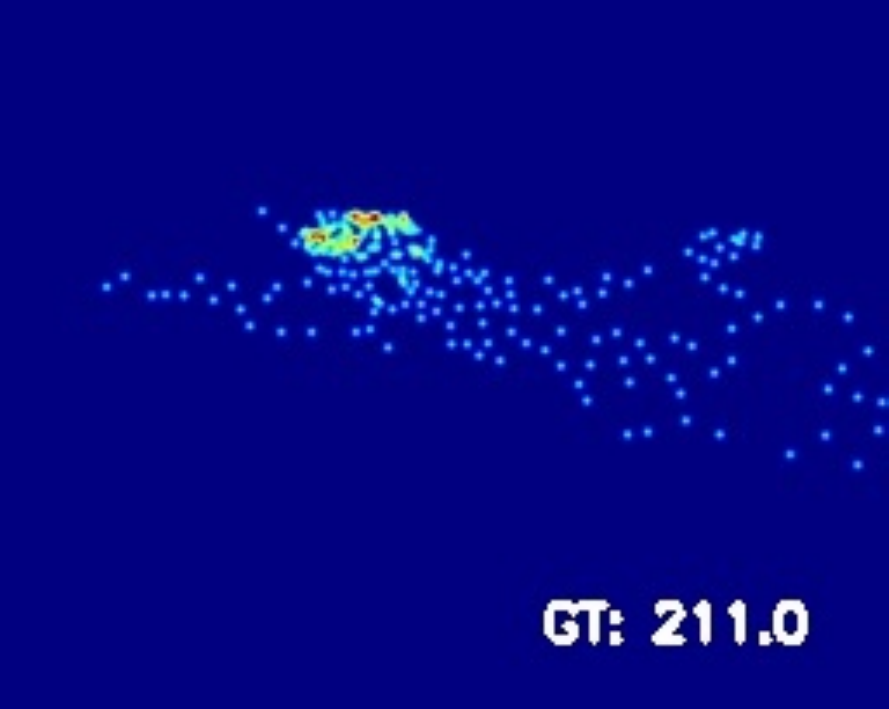}
  \includegraphics[width=2.7cm]{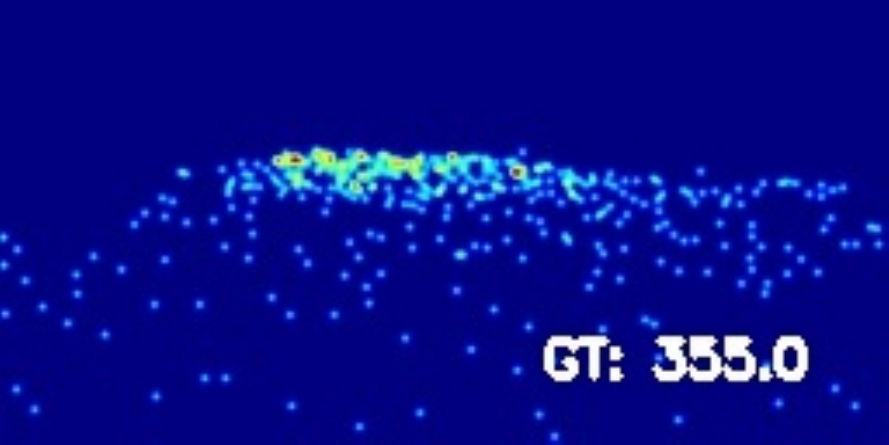}
  \end{minipage}
  }
  \caption{Quantitative results. From left to right: RGB images, CSRNet~\cite{li2018csrnet}, IFNet, ground truth}
  \label{fig:8}
\end{figure}

\begin{table}
\begin{center}
\caption{Compare the transferability of different methods between different datasets}
\label{table:transferability}
\scriptsize
\begin{tabular}{|c|c|c|c|c|c|c|c|c|}
\hline
\multirow{2}*{Method}  & \multicolumn{2}{c|}{A$\rightarrow$B} & \multicolumn{2}{c|}{B$\rightarrow$A} & \multicolumn{2}{c|}{A$\rightarrow$UCF-QNRF} & \multicolumn{2}{c|}{UCF-QNRF$\rightarrow$A} \\ \cline{2-9} & MAE & MSE & MAE & MSE & MAE & MSE & MAE & MSE \\
\hline
\noalign{\smallskip}
\hline
MCNN~\cite{zhang2016single}  & 85.2 & 142.3 & 221.4 & 357.8 & - &  - & - & - \\
D-ConvNet-v1~\cite{shi2018crowd} & 49.1 & 99.2 & 140.4 & 226.1 & - &  - & - & - \\
SPN+L2SM~\cite{xu2019learn} & 21.2 & 38.7 & \textbf{126.8} & \textbf{203.9} & 227.2 & 405.2 & 73.4 & 119.4 \\
\hline
\noalign{\smallskip}
\hline
\textbf{IFNet (ours)} & \textbf{18.4} & \textbf{25.4} & 160.1 & 266.2 & \textbf{201.7} & \textbf{401.8} & \textbf{70.3} & \textbf{113.1} \\
\hline
\end{tabular}
\end{center}
\end{table}
\subsection{Comparison to other methods}
In practical applications, we also hope to get high-quality density maps. Therefore, we use PSNR and SSIM as measurements and compare with previous methods on several datasets. Table 4 shows the results of the experiment. It can be found that the quality of the density map generated by IFNet achieves the best performance . This shows that IFNet has the ability of extracting the spatial features of the crowd in the input images, which benefits from the additional location information provided by the segmentation task.

Another key issue in crowd counting is the generalization ability. Because these datasets contain a small amount of data and are collected from different scenes. This means the model's performance is difficult to transfer to new dataset. But the transferability is significant to the application. Therefore, we test the transferability of the model between different datasets. We train the model on the source dataset and test it on the target dataset. The test results are shown in Table 5. A represents ShanghaiTech Part A dataset and B represents ShanghaiTech Part B dataset. It demonstrates our model could get the state-of-the-art results on several datasets.
\begin{table}
\begin{center}
\caption{Quantitative results of state-of-the-art algorithms}
\label{table:Quantitative results}
\scriptsize
\begin{tabular}{|c|c|c|c|c|c|c|c|c|}
\hline
\multirow{2}*{Method}  & \multicolumn{2}{c|}{ShanghaiTech Part A} & \multicolumn{2}{c|}{ShanghaiTech Part B} & \multicolumn{2}{c|}{UCF\_CC\_50} & \multicolumn{2}{c|}{UCF-QNRF} \\ \cline{2-9} & MAE & MSE & MAE & MSE & MAE & MSE & MAE & MSE \\
\hline\hline
MCNN~\cite{zhang2016single}  & 110.2 & 173.2 & 26.4 & 41.3 & 377.6 & 509.1 & 277 & 426 \\
Cascaded-MTL~\cite{sindagi2017cnn} & 101.3 & 152.4 & 20.0 & 31.1 & 322.8 & 397.9 & 252 & 514 \\
CP-CNN~\cite{sindagi2017generating} & 73.6 & 106.4 & 20.1 & 30.1 & 295.8 & 320.9 & - & - \\
CSRNet~\cite{li2018csrnet} & 68.2 & 115.0 & 10.6 & 16.0 & 266.1 & 397.5 & - & - \\
RAZNet~\cite{liu2019recurrent} & 65.1 & 106.7 & 8.4 & 14.1 & - & - & 116 & 195 \\
ANF~\cite{zhang2019attentional} & 63.9 & 99.4 & 8.3 & 13.2 & 250.2 & 340.0 & 110 & 174\\
SPN+L2SM~\cite{xu2019learn} & 64.2 & \textbf{98.4} & \textbf{7.2} & \textbf{11.1} & \textbf{188.4} & 315.3 & 104.7 & 173.6\\
\hline\hline
\textbf{IFNet (ours)} & \textbf{61.1} & 111.7 & 7.6 & 12.1 & 193.8 & \textbf{272.3} & \textbf{98.2} & 184.6 \\
\hline
\end{tabular}
\end{center}
\end{table}

We compare IFNet and previous methods on several benchmark datasets in Table 6 and show some samples in Figure 5. It indicates that our model achieves comparable performance with the state-of-the-art models on these benchmark datasets.

\section{Conclusions}
In this paper, we propose the Information Fusion Module (IFM) for the flow of information between different columns, which reduces the burden of information redundancy in the network. Therefore IFNet can better capture the various scale features of the image. We also introduce the segmentation task to assist the counting task. However, due to the contradiction between the two tasks, segmentation cannot have a positive impact. We solve this problem by introducing Steeper Differentiable Binarization (SDB). This enables segmentation branch to provide crowd location information to the model. In the experiments, LFNet shows its generalization ability on several datasets.
\clearpage

\bibliography{egbib}

\end{document}